\documentclass[letterpaper]{article} 
\usepackage{aaai25} 
\usepackage{times}  
\usepackage{helvet}  
\usepackage{courier}  
\usepackage[hyphens]{url}  
\usepackage{graphicx} 
\usepackage{natbib}  
\usepackage{caption} 
\usepackage{algorithm}
\usepackage{algorithmic}
\usepackage{booktabs}
\usepackage{multirow} 
\usepackage{enumitem}
\usepackage{makecell}
\usepackage{xcolor} 
\usepackage{pifont}
\usepackage{amsmath}
\usepackage{amssymb}
\usepackage{rotating}
\usepackage{algorithm}
\usepackage{algorithmic}
\usepackage{mathrsfs}
\newcommand{\Epsilon}{\mathcal{E}} 
\usepackage{graphicx}
\usepackage{booktabs}
\usepackage[accsupp]{axessibility} 
\usepackage{bbding}
\usepackage{color}
\usepackage{booktabs}
\usepackage{listings}
\usepackage{multirow}
\usepackage{xcolor} 
\usepackage{color, colortbl}
\usepackage{hyperref}
\hypersetup{
    colorlinks=true,    
    linkcolor=black,    
    citecolor=black,    
    filecolor=magenta,  
    urlcolor=blue       
}
\definecolor{mylb}{RGB}{229, 247, 255}

\definecolor{mygray}{gray}{.9}
\newcommand{\answerYes}[1]{\textcolor{blue}{[Yes]}}
\newcommand{\answerPartial}[1]{\textcolor{blue}{[Partial]}}
\newcommand{\answerNo}[1]{\textcolor{red}{[No]}}
\newcommand{\answerNA}[1]{\textcolor{gray}{[NA]}}
 
\usepackage{newfloat}
\usepackage{listings}
\DeclareCaptionStyle{ruled}{labelfont=normalfont,labelsep=colon,strut=off} 
\floatstyle{ruled}
\newfloat{listing}{tb}{lst}{}
\floatname{listing}{Listing}
\title{VTON-HandFit: Virtual Try-on for Arbitrary Hand Pose Guided by \\ Hand Priors Embedding
}

\author {
    Yujie Liang\textsuperscript{\rm 1}\footnotemark[1], 
    Xiaobin Hu\textsuperscript{\rm 2}\footnotemark[1], Boyuan Jiang\textsuperscript{\rm 2}, Donghao Luo\textsuperscript{\rm 2}\footnotemark[2], Kai Wu\textsuperscript{\rm 2}, \\ Wenhui Han\textsuperscript{\rm 2}, Taisong Jin\textsuperscript{\rm 1}\footnotemark[2], Chengjie Wang\textsuperscript{\rm 2} 
} 
\affiliations{
    $^1$Xiamen University, $^2$Tencent\\
}

\usepackage{bibentry}

\begin{document}

\twocolumn[{%
\renewcommand\twocolumn[1][]{#1}%
\maketitle
\begin{center}
    \centering
    \captionsetup{type=figure}
    \includegraphics[width=0.9\textwidth]{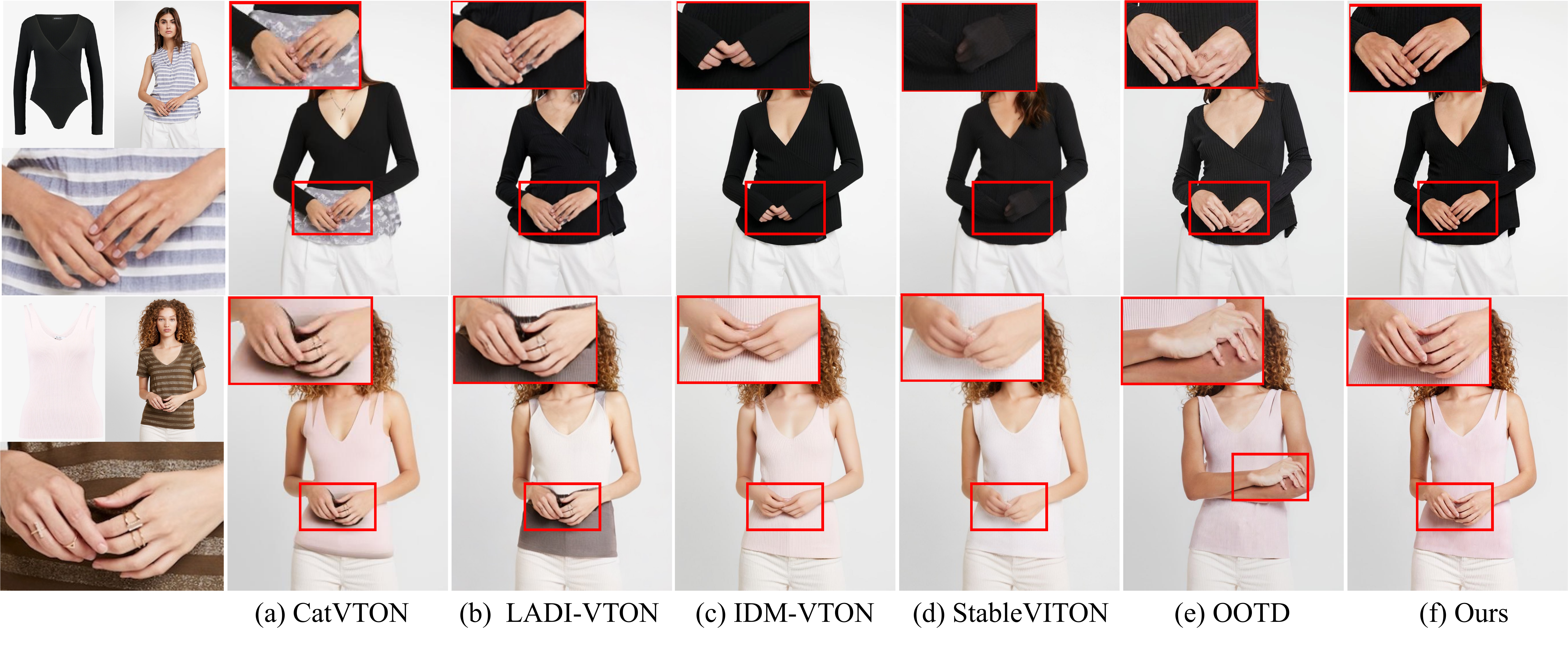}
    \vspace{-15pt}
    \captionof{figure}{\small Comparison of different models on virtual try-on with hand occlusion. The leftmost three images are reference images from the VITON-HD test set, showing the target clothing, the model, and a close-up of the model's hand. Our model excels in preserving hand details and achieving realistic clothing transfer, as highlighted in the red boxes. 
    CatVTON and LADI-VTON utilize a parsing model to retain the hand segment, resulting in the inevitable persistence of residual artifacts and backgrounds from the model image.
    }
    \label{fig:fig1}
\end{center}%
}]

\renewcommand{\thefootnote}{\fnsymbol{footnote}}
\setcounter{footnote}{0} 
\footnotetext[1]{Equal contributions.}
\footnotetext[2]{Corresponding authors.}

\begin{abstract}
Although diffusion-based image virtual try-on has made considerable progress, emerging approaches still struggle to effectively address the issue of hand occlusion (\textit{i.e.,} clothing regions occluded by the hand part), leading to a notable degradation of the try-on performance. 
To tackle this issue widely existing in real-world scenarios, we propose VTON-HandFit, leveraging the power of hand priors to reconstruct the appearance and structure for hand occlusion cases. 
Firstly, we tailor a \textit{Handpose Aggregation Net} using the ControlNet-based structure explicitly and adaptively encoding the global hand and pose priors.
Besides, to fully exploit the hand-related structure and appearance information, we propose \textit{Hand-feature Disentanglement Embedding} module to disentangle the hand priors into the hand structure-parametric and visual-appearance features, and customize a masked cross attention for further decoupled feature embedding.
Lastly, we customize a hand-canny constraint loss to better learn the structure edge knowledge from the hand template of model image.
VTON-HandFit outperforms the baselines in qualitative and quantitative evaluations on the public dataset and our self-collected hand-occlusion Handfit-3K dataset particularly for the arbitrary hand pose occlusion cases in real-world scenarios. The code and dataset will be available at 
\href{https://github.com/VTON-HandFit/VTON-HandFit}{https://github.com/VTON-HandFit/VTON-HandFit}.








\end{abstract}

\section{Introduction}
Image-based virtual try-on (VTON) ~\cite{lee2022high} is a widely adopted and highly promising image synthesis technology in the e-commerce industry. Its primary goal is to enhance the shopping experience for consumers and minimize advertising expenses for clothing merchants. The VTON task involves generating an image of a human model wearing a specific garment. Over the past few years, countless researchers 
\cite{yang2020towards,dong2019towards,ge2021disentangled,davide2022dress} have dedicated significant efforts to achieve more realistic and precise virtual try-on results.

Most recent studies on virtual try-on predominantly utilize generative adversarial networks (GANs) \cite{goodfellow2020generative} or latent diffusion models (LDMs) \cite{rombach2022high} for image synthesis. 
However, traditional GAN-based approaches \cite{han2019clothflow,han2018viton,he2022style} often struggle with accurately generating garment folds, realistic lighting and shadows, and lifelike human body representations. Consequently, more recent research has shifted towards LDM-based methods \cite{zhu2023tryondiffusion}, which effectively improve the authenticity of clothed images.
However, these researches focus on generating more realistic and natural try-on images while keeping the detailed features of garments. 
The hands of try-on models are usually strictly on the sides and garments are not occluded by the hands. Such a strict constraint hinders the virtual try-on paving the way toward real-world applications. For instance, as a highly flexible editing task with a wide range of degrees of freedom, digital humans usually meet the occasions that the garments are occluded by the hands, which severely deteriorates the performance of virtual try-on. 

Motivated by the above challenge, basically, there mainly exist two solutions to tackle this arbitrary hand problem:  \textit{1).} perfect hand parsing model to accurately segment the tiny area among fingers. \textit{2).} involving the hand-reconstruction into the try-on processing. As shown in Fig. \ref{fig:fig1}, CatVTON and LADI-VTON use the parsing model to keep the hand part, leading to unavoidably the remaining artifacts and backgrounds from the model image.
The existing parsing models fail to accurately segment the small and fragmented finger areas, thus we embrace the latter strategy to involve the hand-reconstruction by disentangling the hand into the appearance and structure knowledge processing in the try-on stage. 
Then, we propose a novel diffusion-based virtual try-on method, termed as VTON-HandFit, to generate realistic and natural images for arbitrary hand position or hand-occlusion cases with the aid of hand 3D structure-parametric and visual-appearance priors. 
To embed the pose- and hand-related knowledge, we build a Handpose-Net to exploit the body and hand semantic and structure 
information by getting the body (\textit{e.g.,} DWPose, DensPose)
and 3D hand HaMeR-based \cite{zhu2023tryondiffusion} reconstruction information. 
Secondly, considering the over-small character of hand that is easily ignored after down-sampling operation, we use the DINO \cite{zhang2022dino} to dig the local appearance and texture knowledge of hand as the hand-visual embedding features. 
For the structure of hand, we adopt these structure-parametric priors (\textit{e.g.,} 3D MANO-related parameters, handtype and hand-box coordinates) as hand-structure embedding features. 
After getting structure-parametric and visual-appearance features, 
we aim to enhance the model’s ability to respectively learn hand-related patterns.
To achieve this, we inject the hand-structure embedding into the HandPose-Net with a primary focus on enhancing the latent features related to hand structure, and the hand-visual embedding into the mainstream U-Net specifically aiming to enhance texture-related features, respectively. 
The disentanglement learning of structure and visual appearance enhances the network's ability to efficiently and respectively acquire hand-related knowledge. 
To prevent the interference of hand-related priors with garment features, we impose a constraint that restricts the hand-related features to operate solely within the hand mask area of the cross-attention mechanism. Finally, we tailor a hand-canny constraint loss to more effectively acquire the structural edge knowledge from the hand template of the model image.
Our contributions are summarized as follows:
\vspace{-1mm}
\begin{itemize}[topsep=3pt, partopsep=3pt,leftmargin=5pt, itemsep=3pt]
    \item We propose VTON-HandFit, a novel virtual try-on architecture considering to solve the Hand pose occlusion problem popularly existing in real-world scenarios.
    \vspace{-2mm}
    \item We propose a Hand-Pose Aggregation Net to encode the hand and pose structure priors and also adaptively to adjust the weight of each component. Meanwhile, hand-canny constraint loss is designed to learn the hand structural-edge knowledge of model images. 
    \vspace{-2mm}
    \item We propose a novel hand-feature disentanglement embedding module disentangling the hand knowledge into the structure-parametric priors (\textit{e.g.,} MANO parameters) and visual-appearance priors (\textit{e.g.,} color). A mask cross-attention mechanism is proposed to constrain the hand-related features only working on the hand mask area to avoid the garment feature disturbance.
\vspace{-2mm}
    \item We self-collect Handfit-3K test dataset from online retail sites to evaluate hand occlusion performance. Extensive qualitative and quantitative evaluations demonstrate our superiority over state-of-the-art virtual try-on models especially for hand pose occlusion cases, paving the road towards more complicated virtual try-on in real-world.
\end{itemize}





%

\section{Related Works}
\noindent\textbf{Image-based virtual try-on.}
In the context of image-based virtual try-on, the objective is to generate a realistic representation of a target person wearing a specific garment, given a pair of corresponding images. A series of studies \cite{lee2022high, men2020controllable,xie2023gp,yang2023occlumix}
have been conducted using Generative Adversarial Networks (GANs) \cite{goodfellow2020generative}. These studies typically involve the initial deformation of the garment to match the body shape of the person, followed by the utilization of a generator to superimpose the deformed garment onto the person image. Despite efforts made to minimize the discrepancy between the warped garment and the person 
\cite{ge2021parser,issenhuth2020not,lee2022high}
, these approaches often lack the ability to generalize well to diverse person images, particularly those with complex backgrounds or intricate poses. There exists GAN-based try-on method \cite{lee2022high} considers occluded problems and designs the warping operation to avoid the garment warping misalignment, which is totally different from diffusion-based algorithms as an inpainting problem.

The remarkable generative capabilities of diffusion models have sparked interest in incorporating them into fashion synthesis, particularly in tasks like visual try-on 
\cite{bhunia2023person,cao2023difffashion,karras2023dreampose,chen2024wear,li2024anyfit,zhang2024mmtryon,sun2024outfitanyone}.
TryOnDiffusion \cite{zhu2023tryondiffusion} employs Parallel-UNets to preserve garment details and warp the garment for the try-on process. 
Subsequent studies have treated virtual try-on as an exemplar-based image inpainting problem \cite{yang2023paint}. These studies involved fine-tuning the inpainting diffusion models using virtual try-on datasets 
\cite{kim2024stableviton,morelli2023ladi}
to generate virtual try-on images of superior quality. For instance, such as LADI-VTON \cite{morelli2023ladi} and DCI-VTON \cite{gou2023taming} have emerged, treating clothing as pseudo-words or employing warping networks to integrate garments into pre-trained diffusion models. 
IDM-VTON \cite{choi2024improving} incorporates sophisticated attention modules to encode the high-level semantics of the garment and extract low-level features, thereby preserving fine-grained details. 
While these diffusion methods address issues related to the nature and realism of synthesized images, they struggle to meet hard occasions of the hand-based occlusion which very widely exists in the real-world virtual try-on scenarios. 


\noindent\textbf{Conditional control of diffusion models.}
In recent years, latent diffusion models \cite{rombach2022high} have demonstrated significant success in tasks such as text-to-image generation \cite{ruiz2023dreambooth,kumari2023multi,podell2023sdxl,hu2024diffumatting}
and image-to-image generation \cite{kawar2023imagic, saharia2022palette}
. To achieve more controllable generated results, several studies have introduced conditional control to diffusion models 
\cite{ye2023ip,zhao2024uni}. ControlNet \cite{zhang2023adding} adds spatial conditioning controls (\textit{e.g.,} anny edges, Hough lines, user scribbles, human key points) to pretrained diffusion models via zero convolutions manner. 
T2I-Adapter \cite{mou2024t2i} aligns the internal knowledge within T2I models with external control signals, allowing it to be trained based on different conditions. This approach enables the generation of results with rich control and editing effects, particularly in terms of color and structure. 
In our paper, we focus on the hard occasions of hand-based occlusion in image-based VTON task, and tailor a Hand-Pose Aggregation Net to control the body pose and hand structure. Furthermore, the disengagement knowledge of hand structure and appearance embedding is studied and then these embeddings are respectively injected into the attention mechanism, acting as a mask-based regional control.


\noindent\textbf{Reliable human hand generation.}
Previous studies \cite{samuel2024generating} have acknowledged the problem of misshaped hands that are generated by existing text-to-image diffusion models. 
Controlling the hand is considerably more challenging compared to controlling the body pose, primarily due to the relative size of the main subject.
Besides, diffusion models exhibit significantly improved capability in generating plausible whole-body poses.
This difficulty arises from the intricate nature of learning the physical structure and pose of hands from training images, which encompasses extensive deformations and occlusions. 
Handrefiner \cite{lu2023handrefiner} and RHanDS \cite{wang2024rhands} are post-processing solutions adopting a hand mesh reconstruction model to rectify malformed hands. 
VTON-HandFit involves the hand corrections into the generation process not as a post-processing and disentangles the hand priors into the structure and appearance feature embedding for virtual try-on arbitrary hand pose in the real-world. 

\begin{figure*}[t!]
\centering
\includegraphics[width=0.85\textwidth]{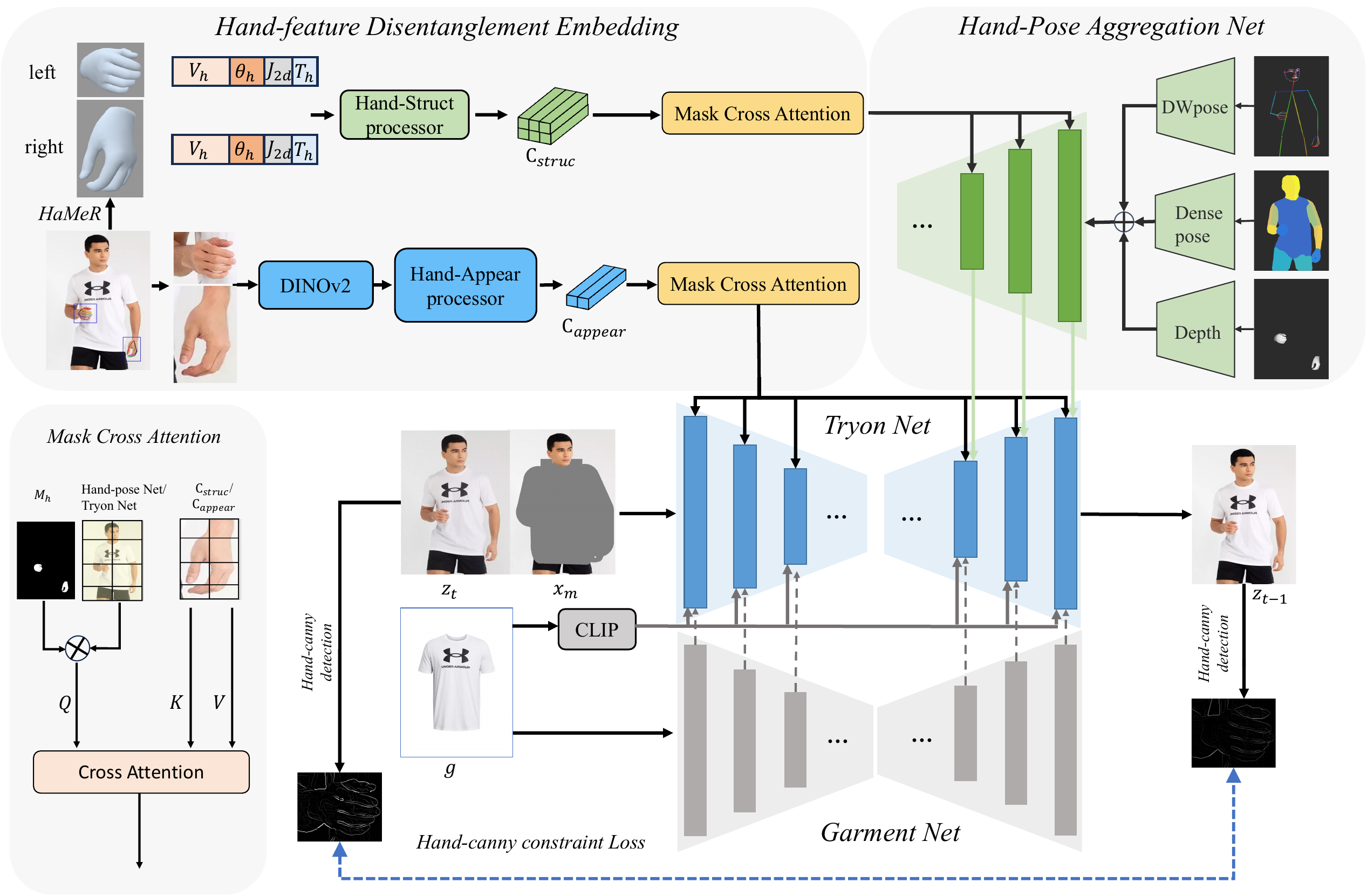}
\vspace{-3mm}
\caption{\small An overview of our VTON-HandFit Network. The network consists of two main components: Hand-feature Disentanglement Embedding and Hand-Pose Aggregation Net. The Hand-feature Disentanglement Embedding module uses the HaMer model to extract hand priors, including hand type ${T_h}$, 3D vertices ${V_h}$, spatial joint locations ${J}_{2d}$, and joint rotation matrices $\theta_h$. These features are processed by the Hand-Struct processor to derive structural features $c_{struc}$. Simultaneously, hand images cropped using bounding boxes are processed by DINOv2 and the Hand-Appear processor to obtain visual features $c_{appear}$. The structural and visual features are integrated using mask cross attention. The Hand-Pose Aggregation Net module controls body and hand poses by aggregating DWpose, Densepose, and hand depth maps. 
}\label{fig:architexture}
\vspace{-6mm}
\end{figure*}

\vspace{-3mm}
\section{Methodology}
\noindent\textbf{Data collection of Handfit-3K.}
Considering that the hand of the current main virtual try-on dataset is separate from the body, there rarely exists a great amount of virtual try-on data for arbitrary hand pose and hard hand-occlusion cases. To meet this urgent need in real-world, we aim to construct a dataset of complex hand poses that occlude the garment. 
Specifically, we collect an additional dataset of 26,000 images sourced from online retail sites. These images underwent processing using a parsing model and a hand bounding box via HaMeR model \cite{pavlakos2024reconstructing} to identify instances of occlusion. 
We devote significant manual effort to inspecting the quality of the data, removing any erroneous or flawed entries, thereby ensuring the high quality of the virtual try-on dataset.
As a result, we obtained a specialized hand occlusion test set comprising 3,620 images, termed as Handfit-3K. 
This dataset facilitates a targeted evaluation of our method's ability to effectively handle occlusions. 
\vspace{-2mm}
\subsection{Preliminary}
\noindent\textbf{Latent Diffusion Models}. 
Our VTON-HandFit is an extension of stable diffusion model, which is a probabilistic model specifically designed to estimate a data distribution p(x) by iteratively reducing noise in a normally distributed variable. Given an input image $x_{0}$, the noise estimation process is:
\vspace{-2mm}
\begin{equation}
\vspace{-1mm}
\mathcal{L}_{noise} = \mathbb{E}_{z \sim \Epsilon(x), C, \epsilon \sim \mathcal{N}(0,1), t} \left[||\epsilon - \epsilon_{\theta}(z_t,t,C)||^2_{2} \right],
\end{equation}
where a Variational AutoEncoder $\Epsilon$ compress  input image $x$ into low-dimension latent space $z$. A conditional U-Net denoiser $\epsilon_{\theta}$ aims to estimate noise $\theta$ in  latent space with the aid of timestep $t$, $t$-th noisy latent representation $z_{t}$, and other text-prompt conditions $C$ extracted by text-encoder.
\vspace{-2mm}
\subsection{VTON-HandFit}
\noindent\textbf{Overview.} An overview of our VTON-HandFit is illustrated in Fig. \ref{fig:architexture}. Given a target person image $x\in \mathbb{R}^{H \times W \times 3}$ and an input garment image $g\in \mathbb{R}^{H \times W \times 3}$, 
VTON-HandFit aims to generate a realistic outfitted image $x_{g}\in \mathbb{R}^{H \times W \times 3}$.
The garment-agnostic person representation $x_{m}$ is masked to erase the garment information by the combination of HumanParsing and HumanPose. Thus, this virtual try-on task is transformed into an example-based inpainting problem to recover the missing mask part. For the input of TryonNet, we simply concatenate the noise image ($z_{t}$) and  latent garment-agnostic features $\mathcal{E}(x_{m})$ to avoid the other pose-relevant information disturbances.
In Hand-Pose Aggregation Net, we also aggregate the pose-related knowledge including the DWPose, DensePose and 3D reconstruction depth map derived by the HaMeR mesh model. 
These pose-related priors are detached with TryonNet to explicitly be learned in Hand-Pose Net. DensePose provides more body-semantic information, DWPose contains abundant skeleton structure, and 3D-hand-derived depth maps demonstrate the pixel-level global-aligned hand position and structure knowledge. These modules complement each other by mutually providing information, and working together to constrain the pose and hand generation. 
In hand-feature disentanglement embedding, we use the HaMeR model for 3D hand reconstruction and get the structure-parametric priors. These features are processed by the Hand-Struct processor to derive structural features $c_{struc}$. Simultaneously, cropped hand images are processed by DINOv2 and the Hand-Appear processor to obtain visual features $c_{appear}$.
Besides,  GarmentNet aims to extract the garment features into the TryonNet mainly by: \textit{ 1).} encoding the garment image into the latent space $\mathcal{E}_{g}$, \textit{2).} using the CLIP to get the image token features for injection.

\noindent\textbf{Hand-Pose Aggregation Net.} 
To aggregate the hand and pose-relevant priors learning, we tailor a HandposeNet to constrain the semantic and skeleton pose, and global pixel-level hand knowledge. Specifically, dwpose, densepose, and render blocks are designed based on the same zero-multilayer structure with different weights to respectively extract the corresponding features.
Zero-multilayer structure consists of seven stacking of \textit{conv} and SiLU activate function with the zero convolution layer at the last layer. 
\vspace{-2mm}
\begin{equation}
\vspace{-1mm}
\mathcal{F}_{hp}=\mathcal{F}_{dw}+\mathcal{F}_{dp}+\mathcal{F}_{rh} \times w_{\mathrm{hand}},
\end{equation}
where $\mathcal{F}_{dw}$, $\mathcal{F}_{dp}$, $\mathcal{F}_{rh}$ are the DWPose feature, Densepose feature, and hand feature after the Zero-multilayer structure, and then the summation is implemented on the pixel-level with the adjustable weight $w_{\mathrm{hand}}$ of  hand feature $\mathcal{F}_{rh}$. To mitigate the impact of detrimental noise on the hidden states of the neural network layers in the trainable HandposeNet during the initial stages of training, we also adopt the zero convolution learning paradigm of ControlNet.

\noindent\textbf{Hand structural embedding.} 
The HaMeR \cite{pavlakos2024reconstructing} model is utilized to reconstruct various MANO parameters and other hand-related information from 2D images, including hand mask ${M_h} \in \mathbb{R}^{n \times 768 \times 1024 }$, hand types $T_h \in \mathbb{R}^{n \times 1 }$, hand 3D vertices $ {V_h} \in \mathbb{R}^{n \times 778 \times 3 }$ , spatial joint locations $ {J}_{2d} \in \mathbb{R}^{n \times 21 \times 2 }$, and joint rotation matrices $ \theta \in \mathbb{R}^{n \times 16 \times 3 \times 3 }$ where $n$ represents the number of hands.
To accommodate variable numbers of hands as input, we pad the hand parameters to a fixed length $n=2$ by adding padding at the end.  
 For hand type ${T_h}$, we use 0 to denote left hands, 1 for right hands, and -1 as a filler for images without hands. Similar padding is applied to other parameters to ensure alignment and correspondence across all dimensions.
We apply Basis Point Set (BPS) \cite{prokudin2019efficient} for dimensionality reduction of hand vertices and encode joint rotations as 6D vectors. 
And then, we import theses information as well as joint rotations and hand types into the hand-structure processor.
The structure of hand-structure processor consists of two parallel blocks: three stacked layers of linear and ReLU activation ($L_{r}$) and two stacked layers of linear and layer normalization ($L_{n}$).
These streamlined hand vertices and joint locations are then fed into $L_{r}$. Concurrently, joint rotations and hand types are transformed via $L_{n}$. 
The outputs from these processes are aggregated to form the Hand Structural Embedding $c_{struc}$, encapsulating essential geometric and 3D information for precise hand representation.
\vspace{-1mm}
\begin{equation}
\vspace{-1mm}
c_{struc} = [ L_{r}(\boldsymbol{{V_h}}),
L_{r}(\boldsymbol{{J}_{2d}}),L_{n}(\boldsymbol{T_h}),L_{n}(\boldsymbol{\theta_h})].
\end{equation}

\noindent\textbf{Hand appearance embedding.} 
To improve hand visual detail, we extract features $\mathcal{F} \in \mathbb{R}^ { n \times 1536} $ from the hand Bounding Box regions using DINOv2 \cite{oquab2023dinov2}. The Hand-Appear processor(HA) uses a linear layer followed by Layer Normalization \cite{ba2016layer}, culminating in the Hand Appearance Embedding $c_{appear}$. This process ensures the hand's visual fidelity and detail are meticulously preserved in our representation.
\vspace{-1mm}
\begin{equation}
\vspace{-1mm}
c_{appear} = \text{HA}(\mathcal{F}).
\vspace{-1mm}
\end{equation}

\begin{figure*}[t!]
\centering
\includegraphics[width=0.75\textwidth]{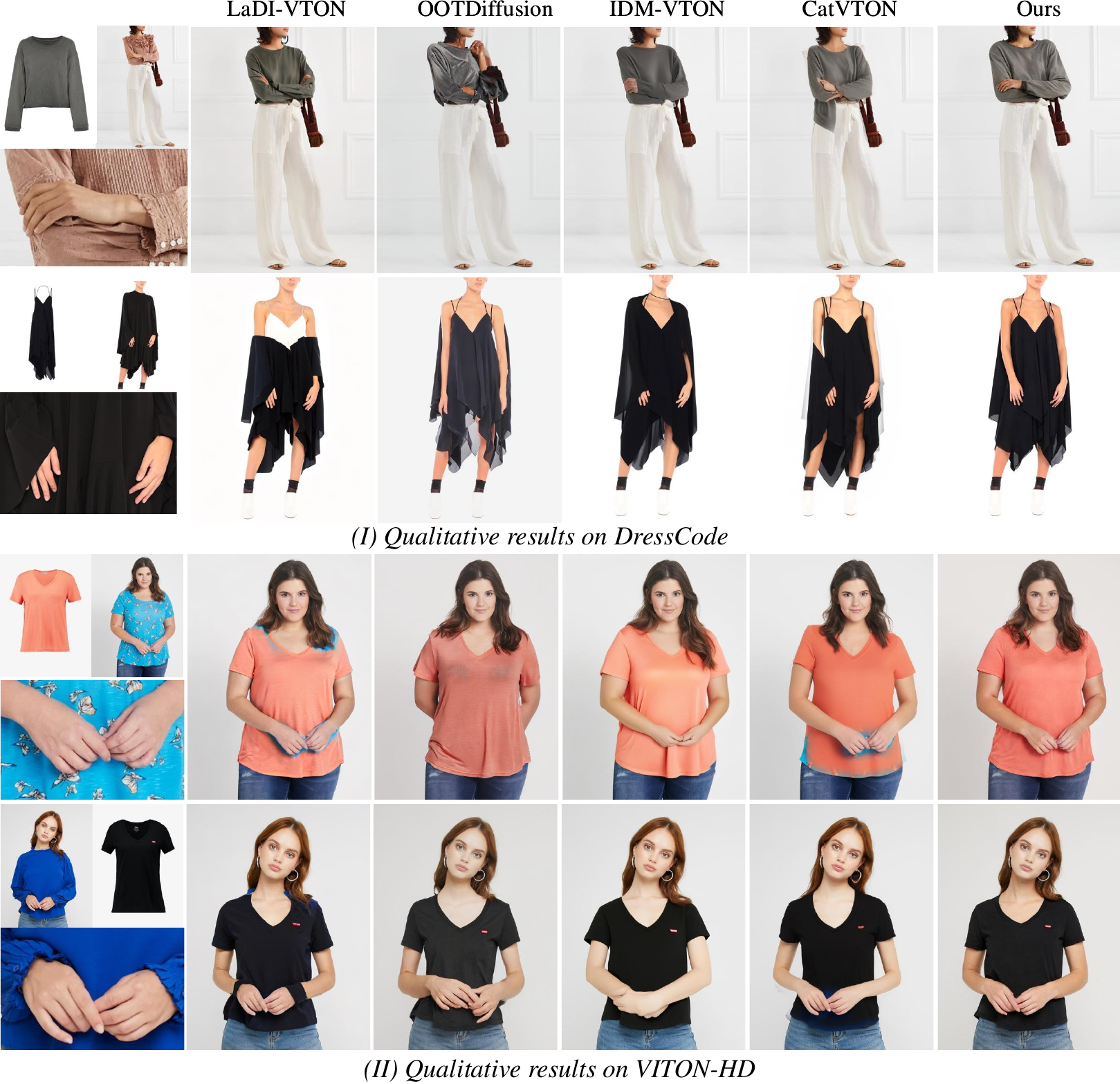}
\vspace{-4mm}
\caption{ \small
Qualitative comparisons of VTON-HandFit with other methods on public datasets: \textit{(I)} Dresscode  and \textit{(II)} VITON-HD. 
}\label{fig:public_dataset}
\vspace{-6mm}
\end{figure*}

\noindent\textbf{Mask-based regional control.} 
Our method introduces a mask-based regional control to enhance hand detail generation and maintain quality in virtual try-on. Utilizing the hand mask ${M_h}$, we modulate Cross-Attention queries, seamlessly integrating Hand Structural Embedding within HandposeNet's Cross-Attention module. Concurrently, Hand Appearance Embedding is adeptly fused into the TryonNet architecture through mask cross-attention, ensuring precise detail preservation.
The mask cross-attention can be represented as:
\vspace{-2mm}
\begin{equation}
\vspace{-1mm}
\begin{split}
Z = \text{Attention}(Q, K, V, M_h) \\
  = \text{Softmax}\left(\frac{QM_hK^\top}{\sqrt{d}}\right)V,
\end{split}
\vspace{-2mm}
\end{equation}
 where $Q$, $K$ and $V$ are the query, key, and value matrices of the attention operation for hand appearance embedding cross-attention, $M_h$ represents the hand mask. 

 
\noindent\textbf{Hand canny shape constraint.} 
During the training process, $\mathcal{L}_{hand}$ is implemented to refine the clarity and precision of hand edges. 
 Specifically, for a small $R_t < T$, we employ one-step denoising process prior to sampling $x_0$ for the direct prediction of $z_0$ from $z_t$ and the model-predicted noise, using Variational Autoencoder for decoder.
 \begin{equation}
z_0 = \frac{z_t - \sqrt{1 - \bar{\alpha}_t} \epsilon_\theta}{\sqrt{\bar{\alpha}_t}}, \quad t \leq R_t.
 \end{equation}
Hand regions are extracted from the generated images using bounding box cropping and are processed with the Canny edge detection operator. These regions are then compared to their corresponding areas in the Ground Truth images using Mean Squared Error (MSE) loss. This loss function is specifically designed to sharpen the model's focus on improving edge details and clarity in the hand regions of the generated images.
\vspace{-2mm}
\begin{equation}
\begin{aligned}
\mathcal{L}_{hand} = \left\{
\begin{array}{ll}
\frac{1}{N} \sum\limits_{i=1}^{N} \left( \phi(I_{gen}^{(i)}) -\phi(I_{GT}^{(i)}) \right)^2 &  t \leq R_t \\
0, &  t > R_t
\end{array},
\right.
\end{aligned}
\vspace{-1mm}
\end{equation}
where $N$ represents the number of images, $I_{gen}^{(i)}$ denotes the $i$-th generated image, and $I_{GT}^{(i)}$ denotes the $i$-th ground truth image. The function $\phi(\cdot)$ applies the Canny edge detection operator to the images.

\begin{figure*}[t!]
\centering
\includegraphics[width=0.75\textwidth]{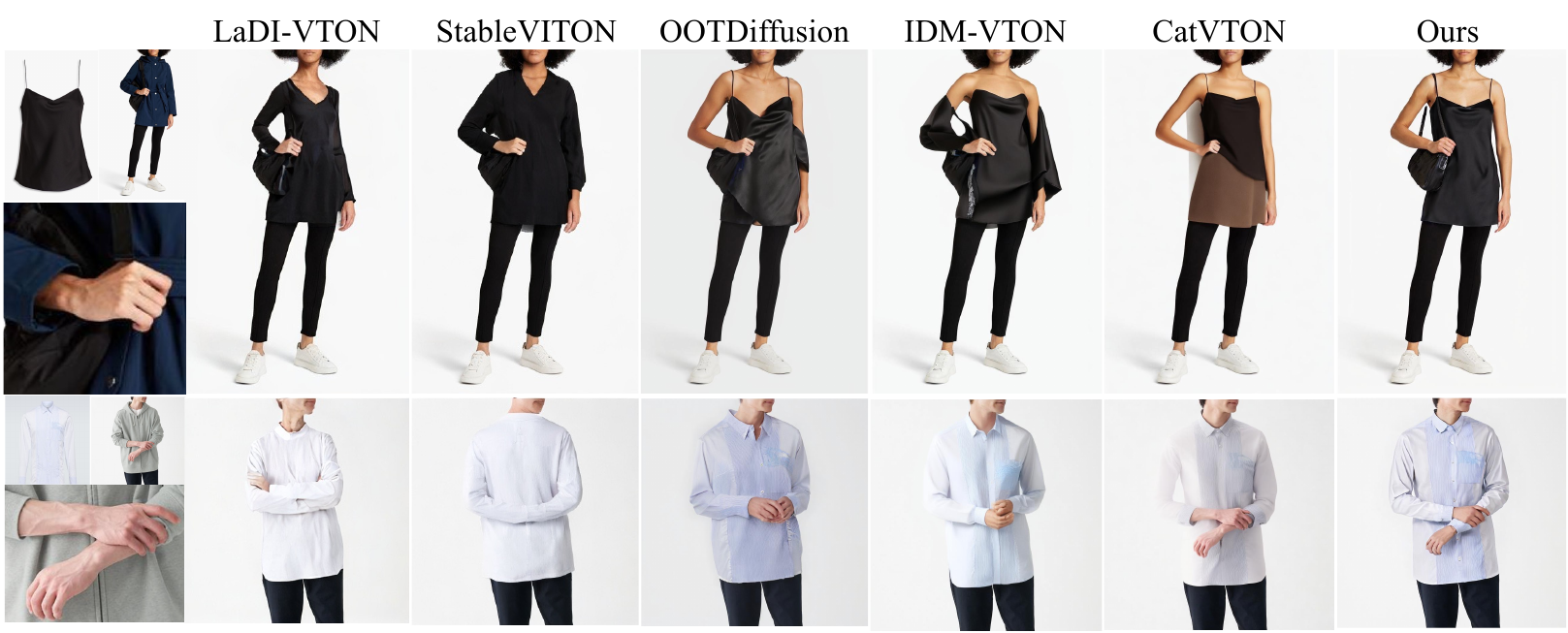}
\vspace{-3mm}
\caption{ \small
Qualitative comparisons of VTON-HandFit with other methods on our Handfit-3K dataset. In Handfit-3K, we remark that the hands in the images are occluded, making it difficult to distinguish them directly through masks.
}\label{fig:testous}
\vspace{-4mm}
\end{figure*}

\begin{table*}[htbp] 
    \centering
    \resizebox{\textwidth}{!}{%
    \begin{tabular}{lcccccccccccc}
        \toprule 
        \multirow{2}{*}{Methods} & \multicolumn{6}{c}{DressCode} & \multicolumn{6}{c}{VITON-HD} \\
        \cmidrule(lr){2-7} \cmidrule(lr){8-13}
        & \multicolumn{4}{c}{Paired} & \multicolumn{2}{c}{Unpaired} & \multicolumn{4}{c}{Paired} & \multicolumn{2}{c}{Unpaired} \\
        \cmidrule(lr){2-5} \cmidrule(lr){6-7} \cmidrule(lr){8-11} \cmidrule(lr){12-13}
        & SSIM $\uparrow$ & LPIPS $\downarrow$ & FID $\downarrow$ & KID $\downarrow$  & FID $\downarrow$ & KID $\downarrow$ & SSIM $\uparrow$ & LPIPS $\downarrow$  & FID $\downarrow$ & KID $\downarrow$ & FID $\downarrow$ & KID $\downarrow$ \\
        \midrule
        LaDI-VTON (2023) & 0.9025 &	0.0719&	4.8636&	1.5580&	6.8421&	2.3345	&	0.8763&	0.0911&	6.6044&	1.0672	&9.4095& 1.6866 \\
        StableVTON (2024) &-	&-&	-	&-	&-	&-		&0.8665	&0.0835	&6.8581&	1.2553&	9.5868	&1.4508 \\

        IDM-VTON (2024) &0.9228&	0.0478	&3.8001	&1.2012	&5.6159	&1.5536		&{0.8806}	&0.0789	&6.3381	& 1.3224	&9.6114	& 1.6387\\
        OOTDiffusion (2024) & 0.8975&	0.0725	&3.9497	&0.7198&	6.7019&	1.8630	&	0.8513&	0.0964&	6.5186&	\textbf{0.8961}	& 9.6733	&1.2061\\
        CatVTON (2024) & 0.9011 &	0.0705 & 3.2755 &\textbf{0.6696} &5.4220 &1.5490   &
        {0.8694} & {0.0970} & {6.1394} & {0.9639} 
        & {9.1434} & {1.2666}\\
        \midrule
        {VTON-HandFit (Ours)} & \textbf{0.9404} & \textbf{0.0444} & \textbf{2.8666} & {0.8009} & \textbf{5.2314} & \textbf{1.0189} & 
        \textbf{0.8850} & \textbf{0.0730} & \textbf{5.9564} & {0.9699} & \textbf{8.8178} & \textbf{0.9951} \\

        \bottomrule
    \end{tabular}
}
\vspace{-3.5mm}
    \caption{\small Quantitative results on VITON-HD and DressCode datasets. We compare the metrics under both paired (model's clothing is the same as the given cloth image) and unpaired settings (model's clothing differs) with other methods. 
    }
    \label{tab:results}
    \vspace{-4mm}
\end{table*}

\begin{table*}[t!] 
    \centering
    \resizebox{\textwidth}{!}{%
    \begin{tabular}{lcccccccccccc}
        \toprule 
        \multirow{2}{*}{Methods}  & \multicolumn{7}{c}{Paired} & \multicolumn{5}{c}{Unpaired} \\
        \cmidrule(lr){2-8} \cmidrule(lr){9-13}
        & SSIM $\uparrow$ & LPIPS $\downarrow$ & FID $\downarrow$ & KID $\downarrow$ & MPJPE$\downarrow$ & FID-H $\downarrow$ & KID-H $\downarrow$ & FID $\downarrow$ & KID $\downarrow$ & MPJPE$\downarrow$ & FID-H $\downarrow$ & KID-H $\downarrow$ \\
        \midrule
        LaDI-VTON (2023) & 0.8597 & 0.1480 & 12.8430 & 4.4112 & 107.6315 & 23.0199 & 17.4558 & 16.2506 & 7.0065 & 111.1547 & 21.1371 & 14.1960 \\
        StableVTON (2024)& 0.8772 & 0.1022 & 11.3675 & 4.1360 & 54.7586 & 22.7470 & 16.699 & 14.0446 & 5.1966 & 57.3334 & 22.8099 & 16.3022 \\
        IDM-VTON (2024)& 0.8816 & 0.0935 & 7.7159 & 2.1287 & 48.6378 & 11.5403 & 3.9727 & 11.2960 & 3.4638 & 57.3447 & 12.7721 & 4.4315 \\
        OOTDiffusion (2024)& 0.8554 & 0.1184 & 8.7104 & 2.1363 & 83.9655 & 12.1984 & 3.6653 & 13.1969 & 4.4839 & 86.1678 & 13.8011 & 4.4479 \\
        CatVTON (2024) & 0.8761 & 0.1123 & 10.6966 & 4.2581 & 60.6608 & 13.1264 & 4.8723 & 15.7081 & 6.9544 & 63.2245 & 15.5603 & 5.6134 \\
        \midrule
        {VTON-HandFit (Ours)} & \textbf{0.8827} & \textbf{0.0819} & \textbf{5.6027} & \textbf{0.4932} & \textbf{25.1376} & \textbf{7.0574} & \textbf{1.2474} & \textbf{10.9786} & \textbf{3.3248} & \textbf{26.6892} & \textbf{9.0003} & \textbf{2.5362} \\
        \bottomrule
    \end{tabular}
    }
    \vspace{-3mm}
    \caption{\small Quantitative results on Handfit-3K dataset in the paired and unpaired settings.}
    \label{tab:results_Handfit}
        \vspace{-7mm}
\end{table*}

\vspace{-2mm}
\section{Experiment}
\noindent\textbf{Baselines.} 
To comprehensively evaluate our approach, we selected five state-of-the-art diffusion-based methods in the virtual try-on domain as baselines for comparison: LaDI-VTON \cite{morelli2023ladi}, StableVTON \cite{kim2024stableviton}, OOTDiffusion \cite{xu2024ootdiffusion}, IDM-VTON \cite{choi2024improving}, and CatVTON \cite{zeng2024cat}. While these methods have made substantial contributions to the virtual try-on field, they exhibit certain limitations in handling hand occlusion issues, which motivated us to propose VTON-HandFit.

\noindent\textbf{Dataset.} 
Our experiments utilize two publicly virtual try-on datasets: VITON-HD \cite{choi2021viton} and DressCode \cite{morelli2022dress}. The VITON-HD dataset consists of 13,679 image pairs, divided into 11,647 for training and 2,032 for testing, featuring front-facing upper-body portraits paired with garments at a resolution of 1024×768. The DressCode dataset, larger and more varied, contains 48,392 training pairs and 5,400 testing pairs, including full-body portraits and a range of garments (upper-body, lower-body, and dresses) at the same resolution. To enhance our evaluation of the model's capability to address hand occlusion challenges, 
we build \textit{Handfit-3K} test with 3,620 images sorted from 26,000 images collected in online retail sites. 




\noindent\textbf{Evaluation metrics.} 
Following previous works \cite{morelli2023ladi,kim2024stableviton}, we adopt both paired and unpaired evaluation settings. In the paired setting, the quality of the generated images is assessed using Structural Similarity (SSIM) \cite{wang2004image}, Learned Perceptual Image Patch Similarity (LPIPS) \cite{zhang2018unreasonable}, Frechet Inception Distance (FID) \cite{heusel2017gans}, and Kernel Inception Distance (KID) \cite{binkowski2018demystifying}. 
For the unpaired setting, we employ FID and KID for comparison. 
Hand generation quality is assessed using FID-H and KID-H for the hand region.
Additionally, we use 2D Mean Per Joint Position Error (MPJPE) \cite{ionescu2013human3} to evaluate hand joint position accuracy. 


\noindent\textbf{Implementation details.} 
Our model is developed using the pre-trained StableDiffusion v1.5 \cite{rombach2022high}. Detailed body dynamics and structural information are captured using DWpose \cite{yang2023effective} for pose estimation, Detectron2 \cite{wu2019detectron2} for Densepose segmentation, and HaMeR for detailed hand depth maps and parameters. 
We use VITON-HD and DressCode datasets for training. 
Our training strategy is implemented in two phases. In the first phase, we train the Hand-Pose Aggregation Net and Hand-feature Disentanglement Embedding modules while freezing the other components. In the second phase, we refine the TryonNet and GarmentNet.
All models are trained uniformly with a batch size of 32, employing the AdamW \cite{loshchilov2017decoupled} optimizer at a learning rate of 5e-5. Training is conducted on 8 NVIDIA V100 GPUs.

\vspace{-2mm}
\subsection{Results on Public Dataset}
\vspace{-2mm}
\noindent\textbf{Qualitative results.} 
Fig. \ref{fig:public_dataset} shows visual comparisons of the VITON-HD and DressCode datasets between VTON-HandFit and other methods. In previous methods, original hand regions are often reused, leading to artifacts such as blue outlines around the hands when garment colors significantly differ (third row). In contrast, our model effectively eliminates these residual backgrounds and preserves natural hand poses, enhancing visual fidelity and pose accuracy. 
 
\noindent\textbf{Quantitative results.} 
Tab. \ref{tab:results} presents the quantitative comparisons of VTON-HandFit against other state-of-the-art methods on VITON-HD and DressCode datasets. By enhancing pose accuracy and the fidelity of hand gestures, our approach achieves markedly improved synthesized image quality. Moreover, 
VTON-HandFit outperforms in FID and KID scores across both datasets in unpaired settings, underscoring the importance of precise hand pose representation for enhancing image quality in practical applications.

\vspace{-3mm}
\subsection{Results on Handfit-3K Dataset}
\noindent\textbf{Qualitative results.} 
Fig. \ref{fig:testous} presents a qualitative comparison between VTON-HandFit and other diffusion-based methods. Our method outperforms in preserving detailed garment attributes and achieving natural hand poses. 
The images generated by VTON-HandFit demonstrate superior consistency in texture, pattern, and fit relative to baseline methods.
Furthermore, the hand poses in our images are more consistent with the original.

\noindent\textbf{Quantitative results.} 
Tab. \ref{tab:results_Handfit} consolidates the performance of VTON-HandFit against contemporary methods on the Handfit-3K dataset. 
VTON-HandFit excels in both paired and unpaired settings, showcasing marked advancements in MPJPE, FID-H, and KID-H metrics. This evidences its capability to generate visually compelling images with precise hand pose representation and detailed hand regions. The model's robustness is further affirmed in the unpaired setting, where it consistently leads across all metrics, reinforcing its practical applicability in virtual try-on scenarios with a focus on hand detail fidelity.

\begin{figure}[t!]
\centering
\includegraphics[width=0.45\textwidth]{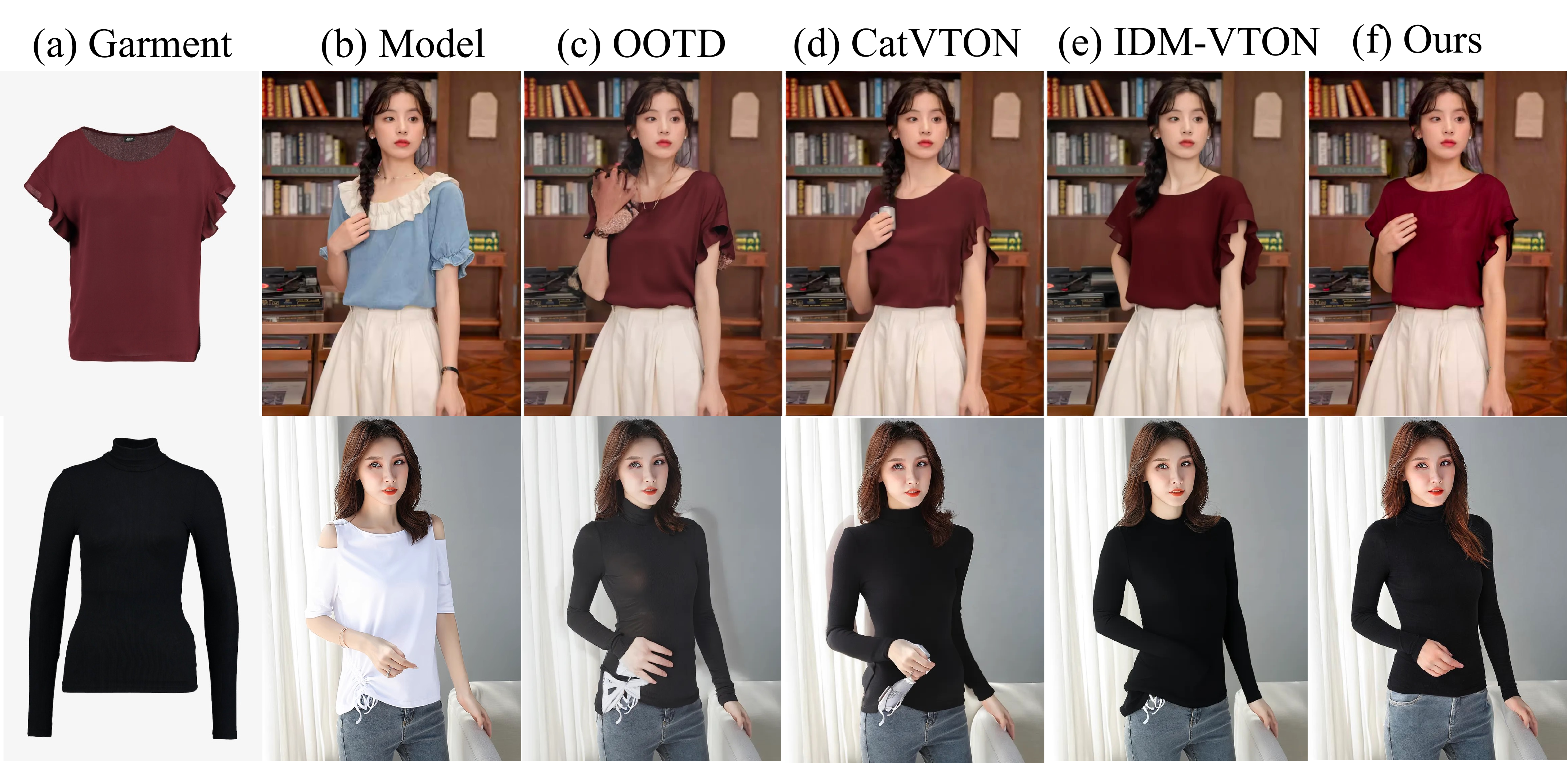}
\vspace{-3mm}
\caption{ \small
Qualitative comparisons in real-world scenarios.
}\label{fig:real-worldxxx}
\vspace{-5mm}
\end{figure}

\noindent\textbf{Results of real-world scenarios.} 
To verify the performance of our VTON-HandFit in real-world scenarios, we collect some cases in the wild  as shown in Fig. \ref{fig:real-worldxxx}. 


\begin{figure}[t!]
\centering
\includegraphics[width=0.45\textwidth]{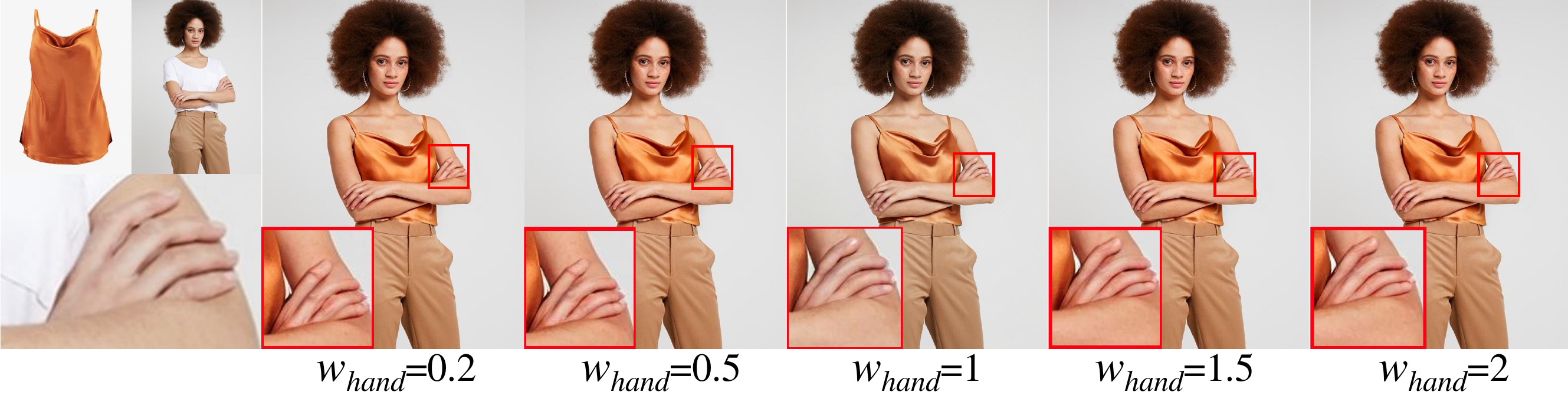}
\vspace{-4mm}
\caption{ \small
Effect of varying control strength $w_{hand}$ on hand shape and texture. As $w_{hand}$ increases from 0.2 to 2, the hand shape and pose better match the depth map, but higher control strengths lead to a loss of hand wrinkles and textures.
}\label{fig:complex_hand_poses}
\vspace{-4mm}
\end{figure}

\begin{figure}[t!]
\centering
\includegraphics[width=0.45\textwidth]{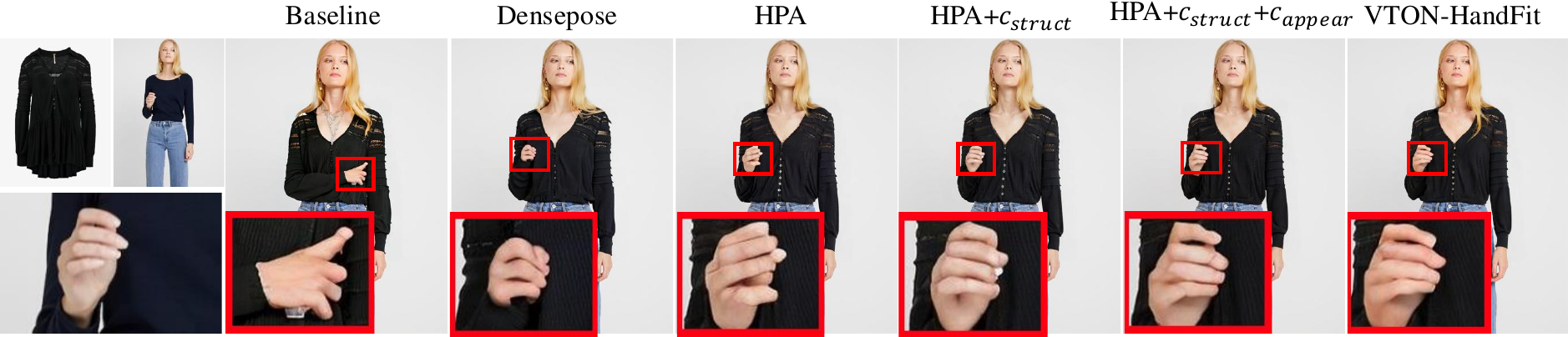}
\vspace{-3mm}
\caption{ \small
Qualitative comparisons of VTON-HandFit with variants on VITON-HD dataset.
}\label{fig:real-world}
\vspace{-2mm}
\end{figure}
\vspace{-2mm}
\subsection{Ablation Study}
\noindent\textbf{Effectiveness of each component.} 
We perform an ablation study to evaluate the contributions of the Hand-Pose Aggregation Net, Hand Structural Embedding $C_{struct}$, and Hand Appearance Embedding $C_{appear}$. To facilitate a clear comparison, we introduce several model variants to ascertain the effectiveness of each component.
The `Baseline' model excludes all specialized components. The `Baseline+HPA (Densepose)' variant incorporates only Densepose for pose control. `Baseline+HPA' utilizes Densepose, Depth, and DWpose for pose control. `Baseline+HPA+$C_{struct}$' excludes the appearance embedding. Lastly, `Baseline+HPA+$C_{struct}$+$C_{appear}$' incorporates all components. These variants do not include hand-canny constraint loss. Tab. \ref{tab:results_ablation} presents a quantitative comparison of our ablation experiments, highlighting the impact of adding different components on the image generation quality.
Fig. \ref{fig:real-world} offers a qualitative analysis, showing the necessity of pose guidance for accurate arm pose replication in the model.
By incorporating stronger hand structure and visual guidance, the generation of hands is significantly improved.

\noindent\textbf{Study of adjustable hand weight.} 
We investigate the influence of varying the control strength of depth maps for different hands on the generated images. Fig. \ref{fig:complex_hand_poses} shows that under complex hand poses, the model may produce deformed hand postures. By increasing the control strength of the depth maps, more accurate and complete hand formations are achieved.

\begin{table}[t!] 
    \centering
    \resizebox{\columnwidth}{!}{%
    \begin{tabular}{lcccc}
        \toprule 
        {Methods} & SSIM $\uparrow$ & LPIPS $\downarrow$ & FID $\downarrow$ & KID $\downarrow$    \\
        \midrule
        Baseline & 0.8619 &0.0923&6.9549&1.2651  \\
        Baseline+HPA(Densepose) & 0.8660 & 0.0876	&6.8924 &1.1254  \\
        Baseline+HPA  &0.8738 & 0.0812 &6.3390 &\textbf{0.9502}   \\
        Baseline+HPA+$C_{struct}$   & 0.8784& 	0.0787& 	6.2225& 	1.1186	 \\
        Baseline+HPA+$C_{struct}+C_{appear}$   &0.8787	&0.0792 &	6.0515 &	1.0084 \\
        VTON-HandFit & \textbf{0.8850} & \textbf{0.0730} & \textbf{5.9564} & {0.9699}  \\
        \bottomrule
    \end{tabular}
    }
    \vspace{-3mm}
    \caption{\small Comparison of different variants on VITON-HD dataset.}
    \label{tab:results_ablation}
    \vspace{-5mm}
\end{table}

\vspace{-2mm}
\section{Conclusion}

We propose VTON-HandFit, a novel model designed for addressing hand occlusion in virtual try-on scenario. 
Specifically, we have developed a Handpose Aggregation Net by explicitly and adaptively encoding global hand and pose priors using the ControlNet-based structure. To fully exploit the hand-related structure and appearance information, we propose the Hand-feature Disentanglement Embedding module, which explicitly separates the hand priors into hand structure-parametric and visual-appearance features. 
Lastly, we have customized a hand-canny constraint loss to enhance the learning of structural edge knowledge from the hand template of the model image. Our approach has been extensively evaluated qualitatively and quantitatively, demonstrating its superiority over state-of-the-art virtual try-on models, particularly in cases involving hand pose occlusion. These results pave the way for more complex virtual try-on applications toward real-world scenarios.

\bibliography{aaai25}

\begin{thebibliography}{58}
\providecommand{\natexlab}[1]{#1}

\bibitem[{Ba, Kiros, and Hinton(2016)}]{ba2016layer}
Ba, J.~L.; Kiros, J.~R.; and Hinton, G.~E. 2016.
\newblock Layer normalization.
\newblock \emph{arXiv preprint arXiv:1607.06450}.

\bibitem[{Bhunia et~al.(2023)Bhunia, Khan, Cholakkal, Anwer, Laaksonen, Shah, and Khan}]{bhunia2023person}
Bhunia, A.~K.; Khan, S.; Cholakkal, H.; Anwer, R.~M.; Laaksonen, J.; Shah, M.; and Khan, F.~S. 2023.
\newblock Person image synthesis via denoising diffusion model.
\newblock In \emph{Proceedings of the IEEE/CVF Conference on Computer Vision and Pattern Recognition}, 5968--5976.

\bibitem[{Bi{\'n}kowski et~al.(2018)Bi{\'n}kowski, Sutherland, Arbel, and Gretton}]{binkowski2018demystifying}
Bi{\'n}kowski, M.; Sutherland, D.~J.; Arbel, M.; and Gretton, A. 2018.
\newblock Demystifying mmd gans.
\newblock \emph{arXiv preprint arXiv:1801.01401}.

\bibitem[{Cao et~al.(2023)Cao, Chai, Hao, Zhang, Chen, and Wang}]{cao2023difffashion}
Cao, S.; Chai, W.; Hao, S.; Zhang, Y.; Chen, H.; and Wang, G. 2023.
\newblock Difffashion: Reference-based fashion design with structure-aware transfer by diffusion models.
\newblock \emph{IEEE Transactions on Multimedia}.

\bibitem[{Chen et~al.(2024)Chen, Chen, Zhai, Ju, Hong, Lan, and Xiao}]{chen2024wear}
Chen, M.; Chen, X.; Zhai, Z.; Ju, C.; Hong, X.; Lan, J.; and Xiao, S. 2024.
\newblock Wear-any-way: Manipulable virtual try-on via sparse correspondence alignment.
\newblock \emph{arXiv preprint arXiv:2403.12965}.

\bibitem[{Choi et~al.(2021)Choi, Park, Lee, and Choo}]{choi2021viton}
Choi, S.; Park, S.; Lee, M.; and Choo, J. 2021.
\newblock Viton-hd: High-resolution virtual try-on via misalignment-aware normalization.
\newblock In \emph{Proceedings of the IEEE/CVF conference on computer vision and pattern recognition}, 14131--14140.

\bibitem[{Choi et~al.(2024)Choi, Kwak, Lee, Choi, and Shin}]{choi2024improving}
Choi, Y.; Kwak, S.; Lee, K.; Choi, H.; and Shin, J. 2024.
\newblock Improving diffusion models for virtual try-on.
\newblock \emph{ECCV}.

\bibitem[{Davide et~al.(2022)Davide, Matteo, Marcella, Federico, Fabio, and Rita}]{davide2022dress}
Davide, M.; Matteo, F.; Marcella, C.; Federico, L.; Fabio, C.; and Rita, C. 2022.
\newblock Dress code: High-resolution multi-category virtual try-on.
\newblock In \emph{Proceedings of the IEEE/CVF Conference on Computer Vision and Pattern Recognition}.

\bibitem[{Dong et~al.(2019)Dong, Liang, Shen, Wang, Lai, Zhu, Hu, and Yin}]{dong2019towards}
Dong, H.; Liang, X.; Shen, X.; Wang, B.; Lai, H.; Zhu, J.; Hu, Z.; and Yin, J. 2019.
\newblock Towards multi-pose guided virtual try-on network.
\newblock In \emph{Proceedings of the IEEE/CVF international conference on computer vision}, 9026--9035.

\bibitem[{Ge et~al.(2021{\natexlab{a}})Ge, Song, Ge, Yang, Liu, and Luo}]{ge2021disentangled}
Ge, C.; Song, Y.; Ge, Y.; Yang, H.; Liu, W.; and Luo, P. 2021{\natexlab{a}}.
\newblock Disentangled cycle consistency for highly-realistic virtual try-on.
\newblock In \emph{Proceedings of the IEEE/CVF conference on computer vision and pattern recognition}, 16928--16937.

\bibitem[{Ge et~al.(2021{\natexlab{b}})Ge, Song, Zhang, Ge, Liu, and Luo}]{ge2021parser}
Ge, Y.; Song, Y.; Zhang, R.; Ge, C.; Liu, W.; and Luo, P. 2021{\natexlab{b}}.
\newblock Parser-free virtual try-on via distilling appearance flows.
\newblock In \emph{Proceedings of the IEEE/CVF conference on computer vision and pattern recognition}, 8485--8493.

\bibitem[{Goodfellow et~al.(2020)Goodfellow, Pouget-Abadie, Mirza, Xu, Warde-Farley, Ozair, Courville, and Bengio}]{goodfellow2020generative}
Goodfellow, I.; Pouget-Abadie, J.; Mirza, M.; Xu, B.; Warde-Farley, D.; Ozair, S.; Courville, A.; and Bengio, Y. 2020.
\newblock Generative adversarial networks.
\newblock \emph{Communications of the ACM}, 63(11): 139--144.

\bibitem[{Gou et~al.(2023)Gou, Sun, Zhang, Si, Qian, and Zhang}]{gou2023taming}
Gou, J.; Sun, S.; Zhang, J.; Si, J.; Qian, C.; and Zhang, L. 2023.
\newblock Taming the power of diffusion models for high-quality virtual try-on with appearance flow.
\newblock In \emph{Proceedings of the 31st ACM International Conference on Multimedia}, 7599--7607.

\bibitem[{Han et~al.(2019)Han, Hu, Huang, and Scott}]{han2019clothflow}
Han, X.; Hu, X.; Huang, W.; and Scott, M.~R. 2019.
\newblock Clothflow: A flow-based model for clothed person generation.
\newblock In \emph{Proceedings of the IEEE/CVF international conference on computer vision}, 10471--10480.

\bibitem[{Han et~al.(2018)Han, Wu, Wu, Yu, and Davis}]{han2018viton}
Han, X.; Wu, Z.; Wu, Z.; Yu, R.; and Davis, L.~S. 2018.
\newblock Viton: An image-based virtual try-on network.
\newblock In \emph{Proceedings of the IEEE conference on computer vision and pattern recognition}, 7543--7552.

\bibitem[{He, Song, and Xiang(2022)}]{he2022style}
He, S.; Song, Y.-Z.; and Xiang, T. 2022.
\newblock Style-based global appearance flow for virtual try-on.
\newblock In \emph{Proceedings of the IEEE/CVF Conference on Computer Vision and Pattern Recognition}, 3470--3479.

\bibitem[{Heusel et~al.(2017)Heusel, Ramsauer, Unterthiner, Nessler, and Hochreiter}]{heusel2017gans}
Heusel, M.; Ramsauer, H.; Unterthiner, T.; Nessler, B.; and Hochreiter, S. 2017.
\newblock Gans trained by a two time-scale update rule converge to a local nash equilibrium.
\newblock \emph{Advances in neural information processing systems}, 30.

\bibitem[{Hu et~al.(2024)Hu, Peng, Luo, Ji, Peng, Jiang, Zhang, Jin, Wang, and Ji}]{hu2024diffumatting}
Hu, X.; Peng, X.; Luo, D.; Ji, X.; Peng, J.; Jiang, Z.; Zhang, J.; Jin, T.; Wang, C.; and Ji, R. 2024.
\newblock DiffuMatting: Synthesizing Arbitrary Objects with Matting-level Annotation.
\newblock \emph{arXiv preprint arXiv:2403.06168}.

\bibitem[{Ionescu et~al.(2013)Ionescu, Papava, Olaru, and Sminchisescu}]{ionescu2013human3}
Ionescu, C.; Papava, D.; Olaru, V.; and Sminchisescu, C. 2013.
\newblock Human3. 6m: Large scale datasets and predictive methods for 3d human sensing in natural environments.
\newblock \emph{IEEE transactions on pattern analysis and machine intelligence}, 36(7): 1325--1339.

\bibitem[{Issenhuth, Mary, and Calauzenes(2020)}]{issenhuth2020not}
Issenhuth, T.; Mary, J.; and Calauzenes, C. 2020.
\newblock Do not mask what you do not need to mask: a parser-free virtual try-on.
\newblock In \emph{Computer Vision--ECCV 2020: 16th European Conference, Glasgow, UK, August 23--28, 2020, Proceedings, Part XX 16}, 619--635. Springer.

\bibitem[{Karras et~al.(2023)Karras, Holynski, Wang, and Kemelmacher-Shlizerman}]{karras2023dreampose}
Karras, J.; Holynski, A.; Wang, T.-C.; and Kemelmacher-Shlizerman, I. 2023.
\newblock Dreampose: Fashion image-to-video synthesis via stable diffusion.
\newblock In \emph{2023 IEEE/CVF International Conference on Computer Vision (ICCV)}, 22623--22633. IEEE.

\bibitem[{Kawar et~al.(2023)Kawar, Zada, Lang, Tov, Chang, Dekel, Mosseri, and Irani}]{kawar2023imagic}
Kawar, B.; Zada, S.; Lang, O.; Tov, O.; Chang, H.; Dekel, T.; Mosseri, I.; and Irani, M. 2023.
\newblock Imagic: Text-based real image editing with diffusion models.
\newblock In \emph{Proceedings of the IEEE/CVF Conference on Computer Vision and Pattern Recognition}, 6007--6017.

\bibitem[{Kim et~al.(2024)Kim, Gu, Park, Park, and Choo}]{kim2024stableviton}
Kim, J.; Gu, G.; Park, M.; Park, S.; and Choo, J. 2024.
\newblock Stableviton: Learning semantic correspondence with latent diffusion model for virtual try-on.
\newblock In \emph{Proceedings of the IEEE/CVF Conference on Computer Vision and Pattern Recognition}, 8176--8185.

\bibitem[{Kumari et~al.(2023)Kumari, Zhang, Zhang, Shechtman, and Zhu}]{kumari2023multi}
Kumari, N.; Zhang, B.; Zhang, R.; Shechtman, E.; and Zhu, J.-Y. 2023.
\newblock Multi-concept customization of text-to-image diffusion.
\newblock In \emph{Proceedings of the IEEE/CVF Conference on Computer Vision and Pattern Recognition}, 1931--1941.

\bibitem[{Lee et~al.(2022)Lee, Gu, Park, Choi, and Choo}]{lee2022high}
Lee, S.; Gu, G.; Park, S.; Choi, S.; and Choo, J. 2022.
\newblock High-resolution virtual try-on with misalignment and occlusion-handled conditions.
\newblock In \emph{European Conference on Computer Vision}, 204--219. Springer.

\bibitem[{Li et~al.(2024)Li, Zhou, Shang, Lin, Chen, and Ni}]{li2024anyfit}
Li, Y.; Zhou, H.; Shang, W.; Lin, R.; Chen, X.; and Ni, B. 2024.
\newblock AnyFit: Controllable Virtual Try-on for Any Combination of Attire Across Any Scenario.
\newblock \emph{arXiv preprint arXiv:2405.18172}.

\bibitem[{Loshchilov and Hutter(2017)}]{loshchilov2017decoupled}
Loshchilov, I.; and Hutter, F. 2017.
\newblock Decoupled weight decay regularization.
\newblock \emph{arXiv preprint arXiv:1711.05101}.

\bibitem[{Lu et~al.(2023)Lu, Xu, Zhang, Wang, and Tao}]{lu2023handrefiner}
Lu, W.; Xu, Y.; Zhang, J.; Wang, C.; and Tao, D. 2023.
\newblock Handrefiner: Refining malformed hands in generated images by diffusion-based conditional inpainting.
\newblock In \emph{ACM Multimedia 2024}.

\bibitem[{Men et~al.(2020)Men, Mao, Jiang, Ma, and Lian}]{men2020controllable}
Men, Y.; Mao, Y.; Jiang, Y.; Ma, W.-Y.; and Lian, Z. 2020.
\newblock Controllable person image synthesis with attribute-decomposed gan.
\newblock In \emph{Proceedings of the IEEE/CVF conference on computer vision and pattern recognition}, 5084--5093.

\bibitem[{Morelli et~al.(2023)Morelli, Baldrati, Cartella, Cornia, Bertini, and Cucchiara}]{morelli2023ladi}
Morelli, D.; Baldrati, A.; Cartella, G.; Cornia, M.; Bertini, M.; and Cucchiara, R. 2023.
\newblock Ladi-vton: Latent diffusion textual-inversion enhanced virtual try-on.
\newblock In \emph{Proceedings of the 31st ACM International Conference on Multimedia}, 8580--8589.

\bibitem[{Morelli et~al.(2022)Morelli, Fincato, Cornia, Landi, Cesari, and Cucchiara}]{morelli2022dress}
Morelli, D.; Fincato, M.; Cornia, M.; Landi, F.; Cesari, F.; and Cucchiara, R. 2022.
\newblock Dress code: High-resolution multi-category virtual try-on.
\newblock In \emph{Proceedings of the IEEE/CVF conference on computer vision and pattern recognition}, 2231--2235.

\bibitem[{Mou et~al.(2024)Mou, Wang, Xie, Wu, Zhang, Qi, and Shan}]{mou2024t2i}
Mou, C.; Wang, X.; Xie, L.; Wu, Y.; Zhang, J.; Qi, Z.; and Shan, Y. 2024.
\newblock T2i-adapter: Learning adapters to dig out more controllable ability for text-to-image diffusion models.
\newblock In \emph{Proceedings of the AAAI Conference on Artificial Intelligence}, volume~38, 4296--4304.

\bibitem[{Oquab et~al.(2023)Oquab, Darcet, Moutakanni, Vo, Szafraniec, Khalidov, Fernandez, Haziza, Massa, El-Nouby et~al.}]{oquab2023dinov2}
Oquab, M.; Darcet, T.; Moutakanni, T.; Vo, H.; Szafraniec, M.; Khalidov, V.; Fernandez, P.; Haziza, D.; Massa, F.; El-Nouby, A.; et~al. 2023.
\newblock Dinov2: Learning robust visual features without supervision.
\newblock \emph{arXiv preprint arXiv:2304.07193}.

\bibitem[{Pavlakos et~al.(2024)Pavlakos, Shan, Radosavovic, Kanazawa, Fouhey, and Malik}]{pavlakos2024reconstructing}
Pavlakos, G.; Shan, D.; Radosavovic, I.; Kanazawa, A.; Fouhey, D.; and Malik, J. 2024.
\newblock Reconstructing hands in 3d with transformers.
\newblock In \emph{Proceedings of the IEEE/CVF Conference on Computer Vision and Pattern Recognition}, 9826--9836.

\bibitem[{Podell et~al.(2023)Podell, English, Lacey, Blattmann, Dockhorn, M{\"u}ller, Penna, and Rombach}]{podell2023sdxl}
Podell, D.; English, Z.; Lacey, K.; Blattmann, A.; Dockhorn, T.; M{\"u}ller, J.; Penna, J.; and Rombach, R. 2023.
\newblock Sdxl: Improving latent diffusion models for high-resolution image synthesis.
\newblock \emph{arXiv preprint arXiv:2307.01952}.

\bibitem[{Prokudin, Lassner, and Romero(2019)}]{prokudin2019efficient}
Prokudin, S.; Lassner, C.; and Romero, J. 2019.
\newblock Efficient learning on point clouds with basis point sets.
\newblock In \emph{Proceedings of the IEEE/CVF international conference on computer vision}, 4332--4341.

\bibitem[{Rombach et~al.(2022)Rombach, Blattmann, Lorenz, Esser, and Ommer}]{rombach2022high}
Rombach, R.; Blattmann, A.; Lorenz, D.; Esser, P.; and Ommer, B. 2022.
\newblock High-resolution image synthesis with latent diffusion models.
\newblock In \emph{Proceedings of the IEEE/CVF conference on computer vision and pattern recognition}, 10684--10695.

\bibitem[{Ruiz et~al.(2023)Ruiz, Li, Jampani, Pritch, Rubinstein, and Aberman}]{ruiz2023dreambooth}
Ruiz, N.; Li, Y.; Jampani, V.; Pritch, Y.; Rubinstein, M.; and Aberman, K. 2023.
\newblock Dreambooth: Fine tuning text-to-image diffusion models for subject-driven generation.
\newblock In \emph{Proceedings of the IEEE/CVF conference on computer vision and pattern recognition}, 22500--22510.

\bibitem[{Saharia et~al.(2022)Saharia, Chan, Chang, Lee, Ho, Salimans, Fleet, and Norouzi}]{saharia2022palette}
Saharia, C.; Chan, W.; Chang, H.; Lee, C.; Ho, J.; Salimans, T.; Fleet, D.; and Norouzi, M. 2022.
\newblock Palette: Image-to-image diffusion models.
\newblock In \emph{ACM SIGGRAPH 2022 conference proceedings}, 1--10.

\bibitem[{Samuel et~al.(2024)Samuel, Ben-Ari, Raviv, Darshan, and Chechik}]{samuel2024generating}
Samuel, D.; Ben-Ari, R.; Raviv, S.; Darshan, N.; and Chechik, G. 2024.
\newblock Generating images of rare concepts using pre-trained diffusion models.
\newblock In \emph{Proceedings of the AAAI Conference on Artificial Intelligence}, volume~38, 4695--4703.

\bibitem[{Sun et~al.(2024)Sun, Cao, Wang, Tian, Zhang, Zhuo, Zhang, Bo, Zhou, Zhang et~al.}]{sun2024outfitanyone}
Sun, K.; Cao, J.; Wang, Q.; Tian, L.; Zhang, X.; Zhuo, L.; Zhang, B.; Bo, L.; Zhou, W.; Zhang, W.; et~al. 2024.
\newblock OutfitAnyone: Ultra-high Quality Virtual Try-On for Any Clothing and Any Person.
\newblock \emph{arXiv preprint arXiv:2407.16224}.

\bibitem[{Wang et~al.(2024)Wang, Liu, Zhou, Zeng, Li, Ge et~al.}]{wang2024rhands}
Wang, C.; Liu, P.; Zhou, M.; Zeng, M.; Li, X.; Ge, T.; et~al. 2024.
\newblock RHanDS: Refining Malformed Hands for Generated Images with Decoupled Structure and Style Guidance.
\newblock \emph{arXiv preprint arXiv:2404.13984}.

\bibitem[{Wang et~al.(2004)Wang, Bovik, Sheikh, and Simoncelli}]{wang2004image}
Wang, Z.; Bovik, A.~C.; Sheikh, H.~R.; and Simoncelli, E.~P. 2004.
\newblock Image quality assessment: from error visibility to structural similarity.
\newblock \emph{IEEE transactions on image processing}, 13(4): 600--612.

\bibitem[{Wu et~al.(2019)Wu, Kirillov, Massa, Lo, and Girshick}]{wu2019detectron2}
Wu, Y.; Kirillov, A.; Massa, F.; Lo, W.-Y.; and Girshick, R. 2019.
\newblock Detectron2.
\newblock https://github.com/facebookresearch/detectron2.

\bibitem[{Xie et~al.(2023)Xie, Huang, Dong, Zhao, Dong, Zhang, Zhu, and Liang}]{xie2023gp}
Xie, Z.; Huang, Z.; Dong, X.; Zhao, F.; Dong, H.; Zhang, X.; Zhu, F.; and Liang, X. 2023.
\newblock Gp-vton: Towards general purpose virtual try-on via collaborative local-flow global-parsing learning.
\newblock In \emph{Proceedings of the IEEE/CVF Conference on Computer Vision and Pattern Recognition}, 23550--23559.

\bibitem[{Xu et~al.(2024)Xu, Gu, Chen, and Chen}]{xu2024ootdiffusion}
Xu, Y.; Gu, T.; Chen, W.; and Chen, C. 2024.
\newblock Ootdiffusion: Outfitting fusion based latent diffusion for controllable virtual try-on.
\newblock \emph{arXiv preprint arXiv:2403.01779}.

\bibitem[{Yang et~al.(2023{\natexlab{a}})Yang, Gu, Zhang, Zhang, Chen, Sun, Chen, and Wen}]{yang2023paint}
Yang, B.; Gu, S.; Zhang, B.; Zhang, T.; Chen, X.; Sun, X.; Chen, D.; and Wen, F. 2023{\natexlab{a}}.
\newblock Paint by example: Exemplar-based image editing with diffusion models.
\newblock In \emph{Proceedings of the IEEE/CVF Conference on Computer Vision and Pattern Recognition}, 18381--18391.

\bibitem[{Yang et~al.(2020)Yang, Zhang, Guo, Liu, Zuo, and Luo}]{yang2020towards}
Yang, H.; Zhang, R.; Guo, X.; Liu, W.; Zuo, W.; and Luo, P. 2020.
\newblock Towards photo-realistic virtual try-on by adaptively generating-preserving image content.
\newblock In \emph{Proceedings of the IEEE/CVF conference on computer vision and pattern recognition}, 7850--7859.

\bibitem[{Yang et~al.(2023{\natexlab{b}})Yang, Chen, Shi, Li, Chen, and Lin}]{yang2023occlumix}
Yang, Z.; Chen, J.; Shi, Y.; Li, H.; Chen, T.; and Lin, L. 2023{\natexlab{b}}.
\newblock OccluMix: Towards de-occlusion virtual try-on by semantically-guided mixup.
\newblock \emph{IEEE Transactions on Multimedia}, 25: 1477--1488.

\bibitem[{Yang et~al.(2023{\natexlab{c}})Yang, Zeng, Yuan, and Li}]{yang2023effective}
Yang, Z.; Zeng, A.; Yuan, C.; and Li, Y. 2023{\natexlab{c}}.
\newblock Effective whole-body pose estimation with two-stages distillation.
\newblock In \emph{Proceedings of the IEEE/CVF International Conference on Computer Vision}, 4210--4220.

\bibitem[{Ye et~al.(2023)Ye, Zhang, Liu, Han, and Yang}]{ye2023ip}
Ye, H.; Zhang, J.; Liu, S.; Han, X.; and Yang, W. 2023.
\newblock Ip-adapter: Text compatible image prompt adapter for text-to-image diffusion models.
\newblock \emph{arXiv preprint arXiv:2308.06721}.

\bibitem[{Zeng et~al.(2024)Zeng, Song, Nie, Tian, Wang, and Liu}]{zeng2024cat}
Zeng, J.; Song, D.; Nie, W.; Tian, H.; Wang, T.; and Liu, A.-A. 2024.
\newblock CAT-DM: Controllable Accelerated Virtual Try-on with Diffusion Model.
\newblock In \emph{Proceedings of the IEEE/CVF Conference on Computer Vision and Pattern Recognition}, 8372--8382.

\bibitem[{Zhang et~al.(2023)Zhang, Li, Liu, Zhang, Su, Zhu, Ni, and Shum}]{zhang2022dino}
Zhang, H.; Li, F.; Liu, S.; Zhang, L.; Su, H.; Zhu, J.; Ni, L.~M.; and Shum, H.-Y. 2023.
\newblock Dino: Detr with improved denoising anchor boxes for end-to-end object detection.
\newblock \emph{ICLR}.

\bibitem[{Zhang, Rao, and Agrawala(2023)}]{zhang2023adding}
Zhang, L.; Rao, A.; and Agrawala, M. 2023.
\newblock Adding conditional control to text-to-image diffusion models.
\newblock In \emph{Proceedings of the IEEE/CVF International Conference on Computer Vision}, 3836--3847.

\bibitem[{Zhang et~al.(2018)Zhang, Isola, Efros, Shechtman, and Wang}]{zhang2018unreasonable}
Zhang, R.; Isola, P.; Efros, A.~A.; Shechtman, E.; and Wang, O. 2018.
\newblock The unreasonable effectiveness of deep features as a perceptual metric.
\newblock In \emph{Proceedings of the IEEE conference on computer vision and pattern recognition}, 586--595.

\bibitem[{Zhang et~al.(2024)Zhang, Lin, Li, Luo, Kampffmeyer, Dong, and Liang}]{zhang2024mmtryon}
Zhang, X.; Lin, E.; Li, X.; Luo, Y.; Kampffmeyer, M.; Dong, X.; and Liang, X. 2024.
\newblock MMTryon: Multi-Modal Multi-Reference Control for High-Quality Fashion Generation.
\newblock \emph{arXiv preprint arXiv:2405.00448}.

\bibitem[{Zhao et~al.(2024)Zhao, Chen, Chen, Bao, Hao, Yuan, and Wong}]{zhao2024uni}
Zhao, S.; Chen, D.; Chen, Y.-C.; Bao, J.; Hao, S.; Yuan, L.; and Wong, K.-Y.~K. 2024.
\newblock Uni-controlnet: All-in-one control to text-to-image diffusion models.
\newblock \emph{Advances in Neural Information Processing Systems}, 36.

\bibitem[{Zhu et~al.(2023)Zhu, Yang, Zhu, Reda, Chan, Saharia, Norouzi, and Kemelmacher-Shlizerman}]{zhu2023tryondiffusion}
Zhu, L.; Yang, D.; Zhu, T.; Reda, F.; Chan, W.; Saharia, C.; Norouzi, M.; and Kemelmacher-Shlizerman, I. 2023.
\newblock Tryondiffusion: A tale of two unets.
\newblock In \emph{Proceedings of the IEEE/CVF Conference on Computer Vision and Pattern Recognition}, 4606--4615.

\end{thebibliography}
\appendix

\end{document}


\maketitle

\section{Overview}\label{sec:overview}
In this supplementary document, we provide additional results to complement our main paper. Firstly, we present the inference time of our VTON-HandFit during the testing phase. Secondly, we provide more qualitative comparisons with state-of-the-art models. Lastly, we offer a preview of our Handfit-3K. 

\begin{figure*}[t!]
\centering
\includegraphics[width=1\textwidth]{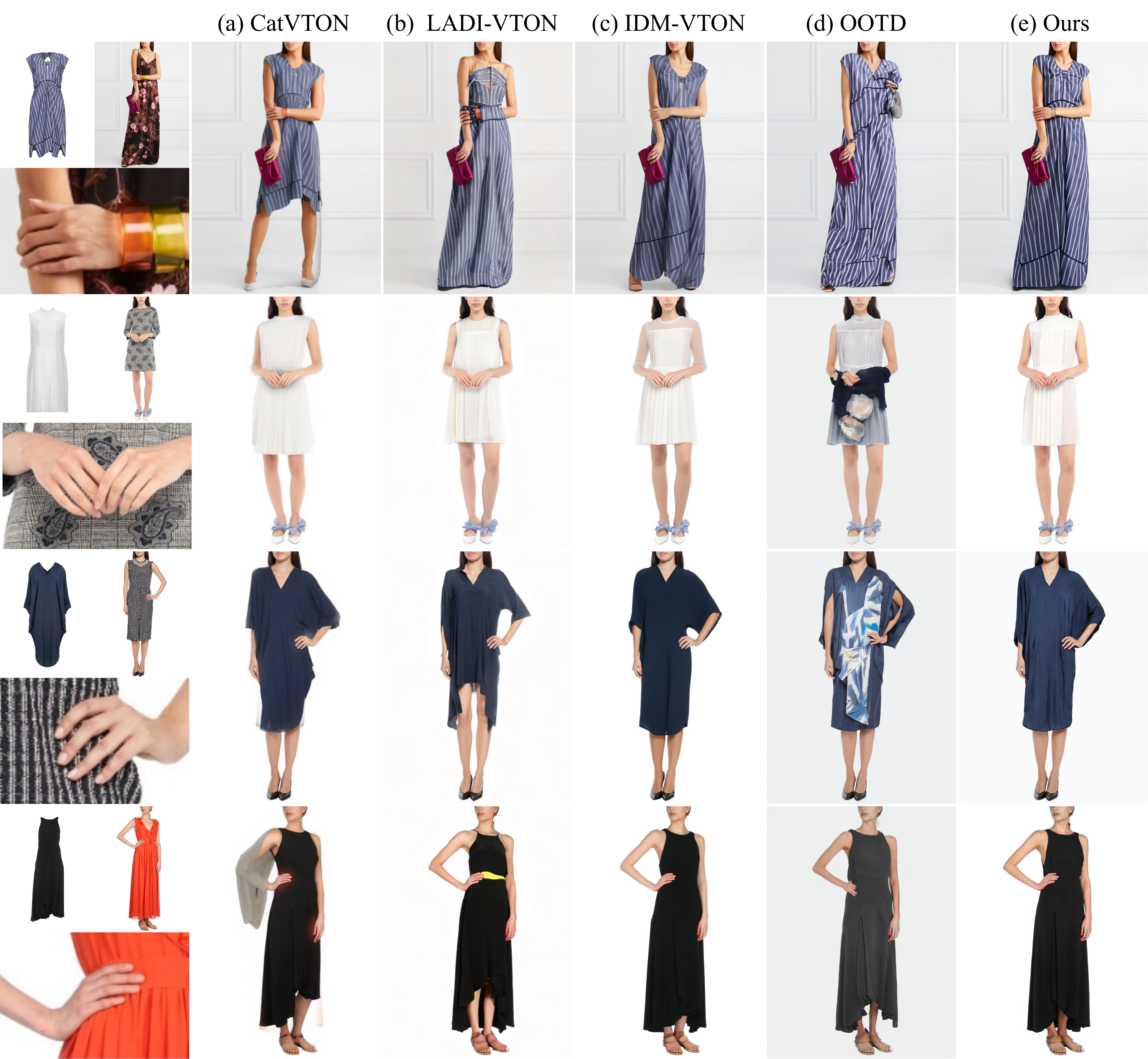}
\caption{ \small 
Qualitative comparisons of VTON-HandFit with other methods on DressCode dataset.
}\label{fig:qualitatin_comparison_dc}
\end{figure*}

\noindent{\textbf{Inference Time.}}
To analyze inference time while excluding I/O operations, we configure the batch size to 1 and set the image resolution at 768$\times$1024. The evaluation is conducted using PyTorch on an Intel(R) Xeon(R) Platinum 8255C CPU @ 2.50GHz and an NVIDIA V100 GPU. We compare our VITON-Handfit model against state-of-the-art methods listed in Tab.  \ref{tab:tab2_inference_speed}.
OOTDiffusion \cite{xu2024ootdiffusion}, IDM-VTON \cite{choi2024improving}, and CatVTON \cite{chong2024catvton} are tested under their default configurations. 
Our approach remains competitive performance across these benchmarks.

\begin{figure*}[t!]
\centering
\includegraphics[width=0.95\textwidth]{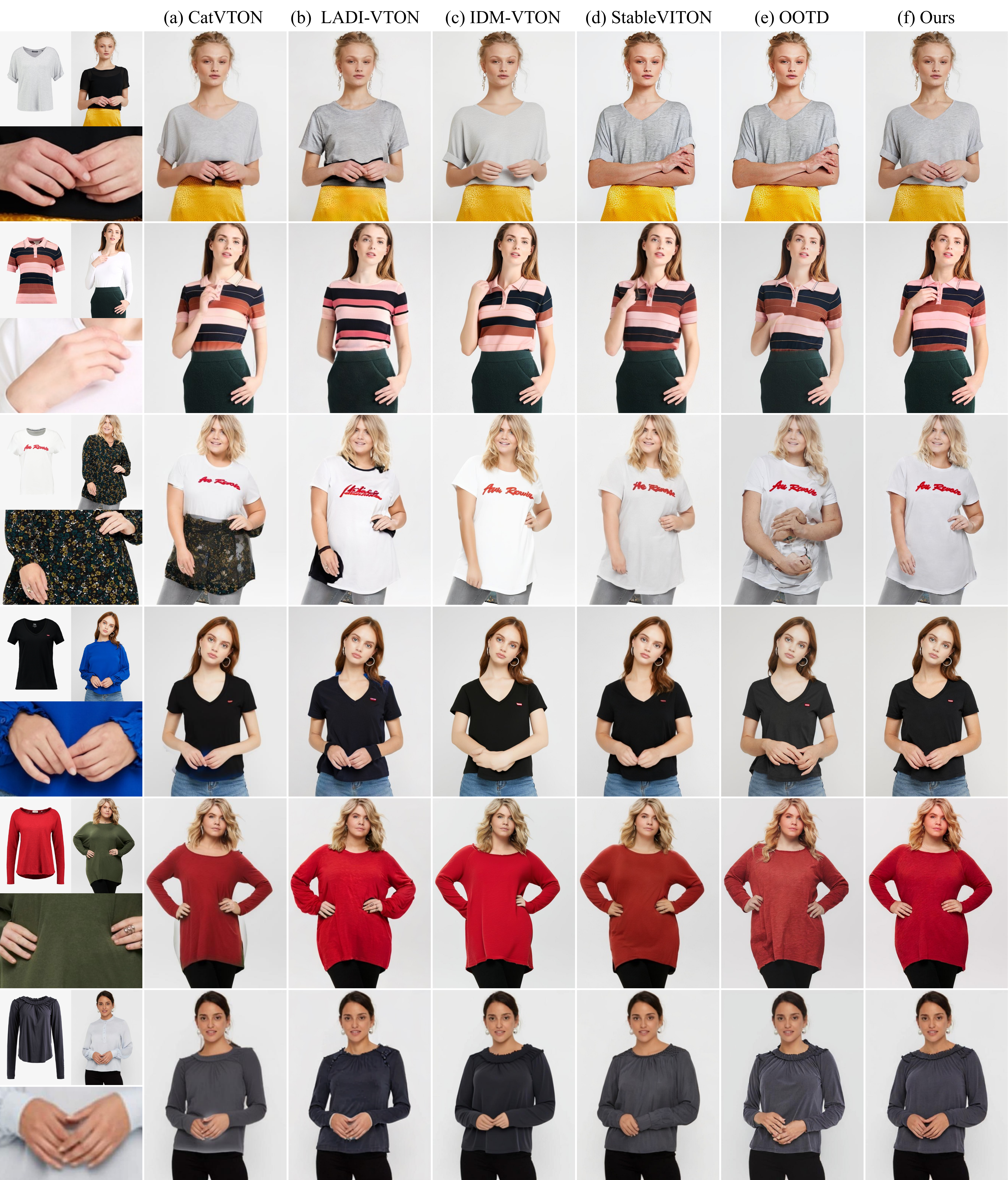}
\caption{ \small
Qualitative comparisons of VTON-HandFit with other methods on VITON-HD dataset.
}\label{fig:qualitatin_comparison_vt}
\end{figure*}

\noindent\textbf{Qualitative Evaluation.}
More qualitative comparisons are presented in Fig. \ref{fig:qualitatin_comparison_dc} for DressCode dataset \cite{morelli2022dress} and in Fig. \ref{fig:qualitatin_comparison_vt} for VITON-HD dataset \cite{choi2021viton}. These comparisons highlight our method's proficiency in generating superior hand poses, especially in scenarios involving hand occlusions.

\begin{figure*}[t!]
\centering
\includegraphics[width=0.95\textwidth]{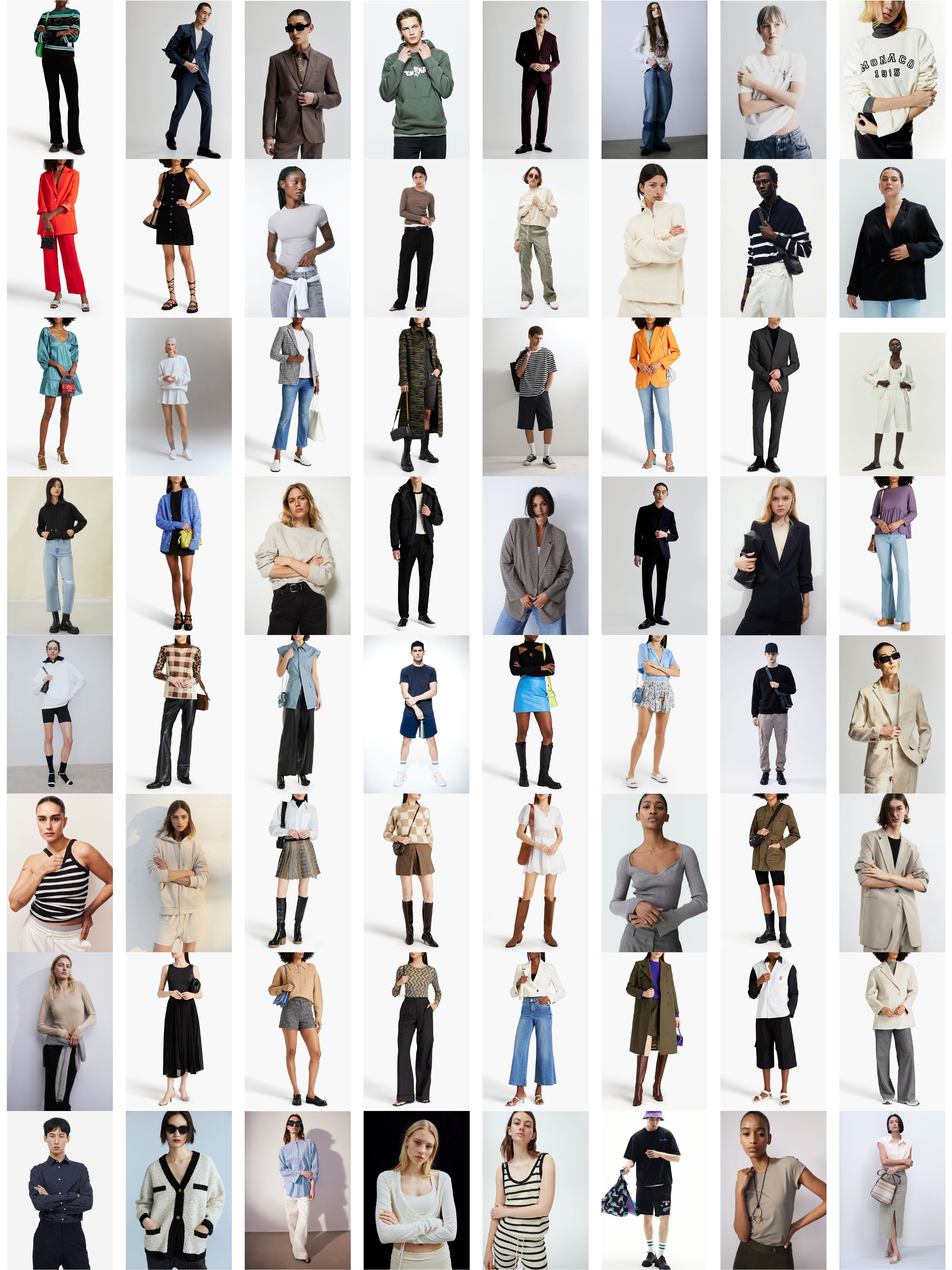}
\caption{ 
A preview of our Handfit-3K dataset. 
}\label{fig:handfit_dataset}
\end{figure*}

\noindent{\textbf{Handfit-3K.}}
We provide additional previews of Handfit-3K images in Fig. \ref{fig:handfit_dataset}. Within the Handfit-3K dataset, hand masks are nearly indiscernible using traditional parsing and OpenPose segmentation method.

\begin{table}[ht!] 
\centering
\footnotesize
\resizebox{\columnwidth}{!}{%
\begin{tabular}{r|c|c|c|c}
\toprule
 & OOTDiffsion & IDM-VTON & CatVTON & Ours \\ 
\hline
Test time (\textit{s}) $\downarrow$ & \textbf{13.332} & 18.258 & 32.462  & \underline{14.075} \\ 
\bottomrule
\end{tabular}
}
\caption{ \small Inference speed (\textit{s}) analyses on VITON-HD dataset. 
The best result is highlighted in \textbf{bold}, while the second-best result is indicated with \underline{underlining}.
}
\label{tab:tab2_inference_speed}
\end{table}

\bibliography{aaai25}